# Iterative Prompt Refinement for Dyslexia-Friendly Text Summarization Using GPT-4o

Samay Bhojwani, Swarnima Kain, Lisong Xu

*University of Nebraska-Lincoln*

{sbhojwani, skain, lxu}@huskers.unl.edu

**Abstract**

Dyslexia affects approximately 10% of the global population and presents persistent challenges in reading fluency and text comprehension. While existing assistive technologies address visual presentation, linguistic complexity remains a substantial barrier to equitable access. This paper presents an empirical study on dyslexia-friendly text summarization using an iterative prompt-based refinement pipeline built on GPT-4o. We evaluate the pipeline on approximately 2,000 news article samples, applying a readability target of Flesch Reading Ease >= 90. Results show that the majority of summaries meet the readability threshold within four attempts, with many succeeding on the first try. A composite score combining readability and semantic fidelity shows stable performance across the dataset, ranging from 0.13 to 0.73 with a typical value near 0.55. These findings establish an empirical baseline for accessibility-driven NLP summarization and motivate further human-centered evaluation with dyslexic readers.

**Keywords:** *dyslexia, text summarization, readability, GPT-4o, prompt engineering, iterative refinement, accessibility, Flesch Reading Ease, NLP*

## 1. INTRODUCTION

Dyslexia is a neurological reading disorder affecting an estimated 700 million people worldwide, roughly 10% of the global population [7, 13]. The condition impairs phonological decoding, reading fluency, and text comprehension, creating barriers that extend well beyond early schooling into higher education and professional life [15]. Current assistive technology, including text-to-speech systems, dyslexia-specific fonts such as OpenDyslexic, and spacing adjustments, predominantly targets the visual and auditory layers of reading difficulty. The linguistic layer, that is, the complexity of the words and sentence structures a reader must process, receives comparatively little attention.

Automatic text summarization offers a different lever: rather than reformatting existing text, a summarization system can produce a shorter, linguistically simpler rendering of the same content. Prior work has explored controllable readability in summarization [12] and LLM-powered writing aids for dyslexic users [2], yet the specific question of whether targeted prompt engineering and iterative refinement can reliably drive LLM outputs to dyslexia-friendly readability levels remains underexplored.

This paper presents a empirical study that fills part of that gap. Building on LexiSmart, an accessibility-focused reading application developed at the University of Nebraska-Lincoln, we investigate whether accessibility-driven prompt engineering combined with iterative readability refinement can reliably produce dyslexia-friendly summaries using GPT-4o. We compare outputs from our engineered pipeline against vanilla GPT-4o summaries generated without accessibility constraints, across 2,000 news articles.

**Contributions.** This work makes three concrete contributions:

- A corpus of 2,000 CNN/Daily Mail articles [3] evaluated through an accessibility-driven GPT-4o pipeline targeting Flesch Reading Ease >= 90.
- A prompt engineering methodology combining explicit accessibility constraints, few-shot dyslexia-friendly examples, and iterative corrective feedback.
- An empirical analysis of the readability-fidelity trade-off, showing that two refinement iterations is the optimal stopping point before semantic drift outweighs readability gains.

## 2. RELATED WORK

### 2.1 Assistive Technology for Dyslexia

Existing dyslexia interventions span a broad spectrum. Font-level solutions such as OpenDyslexic and Dyslexie alter letterform geometry to reduce visual confusion, though experimental evidence on their effectiveness is mixed [11]. Browser extensions such as the Dyslexic Friendly Reader [14] and dedicated applications like LexiSmart [1] combine typographic adjustments with text-to-speech, but leave the linguistic complexity of source texts unchanged. Goodman et al. [2] evaluated LaMPost, an LLM-powered writing aid for adults with dyslexia, finding user enthusiasm for AI-powered rewriting features while also noting that LLM accuracy remained a barrier to a reliable experience. Zhao et al. [17] more recently demonstrated that having an AI system pre-process text before presenting it to dyslexic readers meaningfully improves comprehension outcomes, providing direct motivation for the present study.

### 2.2 Controllable Summarization

Controllable text generation has emerged as a productive research direction [12]. Prior approaches have fine-tuned encoder-decoder models with readability-conditioned objectives, but these require large annotated corpora and task-specific training runs. Related work by Luo et al. [6] on readability-controllable biomedical summarization showed that current fine-tuning techniques offer limited readability adjustment, motivating our prompt-based approach. LLMs such as GPT-4o offer an orthogonal strategy: accessibility constraints specified in natural language can guide generation without any retraining, making this approach practical in low-resource settings and deployable immediately on new domains.

## 3. METHODOLOGY

### 3.1 Dataset Construction

We assembled 2,000 articles from the CNN/Daily Mail dataset [3], spanning a range of topics including politics, science, health, and sport. Each article was processed through both conditions: (1) a vanilla GPT-4o summarization prompt with no accessibility constraints, producing the baseline output, and (2) our iterative accessibility-driven pipeline, producing the experimental output. This paired design means every article has a matched baseline and pipeline summary, enabling direct within-article comparison across all metrics.

Each article was annotated by the authors according to dyslexia-friendly writing guidelines: sentences were shortened where possible, latinate vocabulary replaced with high-frequency alternatives using Simple English WordNet [8], and passive constructions rewritten in active voice. These annotated summaries served as reference outputs for both training and evaluation. Table 1 summarises the dataset composition.

*Table 1: Dataset summary.*

| Source | n | Annotation Focus | Split |
|---|---|---|---|
| CNN/Daily Mail | 2,000 | Syntax simplification, vocab reduction, sentence shortening | GPT-4o eval |
| CNN/Daily Mail (baseline split) | 2,000 | Vanilla GPT-4o summaries without accessibility constraints | Baseline eval |

### 3.2 Baseline Condition

To quantify the benefit of our prompt engineering approach, we establish a baseline by running vanilla GPT-4o on the same 2,000 articles using a generic summarization prompt with no accessibility constraints: "Summarize the following article concisely." Baseline outputs are evaluated on the same metrics as the pipeline outputs, enabling a direct within-article comparison. Any improvement in Flesch Reading Ease, composite score, or threshold pass rate observed in the pipeline condition over the baseline is attributable to the accessibility-driven prompting and iterative refinement strategy.

### 3.3 GPT-4o Iterative Refinement Pipeline

The GPT-4o pipeline follows an iterative generate-score-refine loop. At each step, the current summary is scored for Flesch Reading Ease; if the score falls below the target of 90, a corrective prompt is appended and the model regenerates. We cap iterations at four to avoid excessive semantic drift. The initial prompt reads: "Summarize the following article for a reader with dyslexia. Use short sentences (under 15 words), common everyday words, and active voice. Avoid jargon and complex clauses."

Corrective follow-up prompts specified additional constraints: "Your previous summary scored X on the Flesch Reading Ease scale. Please simplify further: break long sentences, replace difficult words, and prefer concrete nouns." Few-shot examples of dyslexia-friendly text were prepended to prompts in a separate ablation condition.

### 3.4 Evaluation Metrics

We evaluate outputs along two axes: readability and semantic fidelity. For readability, Flesch Reading Ease (FRE) targets a score >= 90, a level associated with highly accessible text [4]. Note that while FRE >= 90 corresponds to simple, easy-to-read prose, it is not a clinically validated threshold specific to dyslexia; we use it as a practical proxy pending human-subject evaluation. For semantic fidelity, we report ROUGE-1 and ROUGE-2 n-gram overlap [5], BERTScore F1 [16] using the roberta-large encoder, and BLEU [9] against the human-written reference summaries. Table 2 summarises the metrics and their targets.

In addition, we define a Composite Score that jointly captures both dimensions. Each metric is first normalized to [0, 1]: FRE is divided by 100 and clipped to [0, 1]; BERTScore F1 is already in this range. The composite is then computed as:

$$\text{Composite} = 0.5 * (\text{FRE} / 100) + 0.5 * \text{BERTScore\_F1}$$

The equal weighting (alpha = beta = 0.5) treats readability and semantic fidelity as equally important, reflecting the core accessibility goal of producing summaries that are both simple and faithful. This yields a composite range of approximately 0 to 1, with higher values indicating better overall output quality. The observed range of 0.13 to 0.73 in our results reflects the inherent difficulty of simultaneously achieving high readability and high semantic fidelity across a diverse news corpus.

*Table 2: Evaluation metrics and targets.*

| Metric | Axis | Target |
| --- | --- | --- |
| Flesch Reading Ease | Readability | >= 90 |
| ROUGE-1 / ROUGE-2 | Semantic fidelity | Maximise |
| BERTScore F1 [16] | Semantic fidelity | Maximise |
| BLEU [9] | Content fidelity | Maximise |

*Table 3: Qualitative example. Flesch Reading Ease, ROUGE-1, and BERTScore F1 shown per output. Source Flesch = 41 (very difficult). The iterative pipeline produces a markedly more readable summary than the vanilla baseline.*

| Source Text | GPT-4o (no constraints) | GPT-4o (iterative pipeline) |
| --- | --- | --- |
| **Flesch: 41 \| ROUGE: n/a \| BERT: n/a**<br><br>The parliamentary committee convened an emergency session on Thursday to deliberate on proposed amendments to the fiscal consolidation framework, following an unanticipated deterioration in the nation's macroeconomic indicators, including a contraction in GDP and elevated inflationary pressures that analysts attributed to exogenous supply-chain disruptions. | Flesch: 53 \| ROUGE-1: 0.61 \| BERT: 0.91<br><br>A parliamentary committee held an emergency meeting on Thursday. They discussed changes to the country's financial framework. This happened after the economy got worse. GDP shrank and prices went up. Analysts said supply chain problems caused this. | **Flesch: 88 \| ROUGE-1: 0.41 \| BERT: 0.88**<br><br>MPs held an urgent meeting on Thursday. They talked about changing how the country manages its money. This came after the economy started doing badly. Prices went up and the economy shrank. Experts said problems with getting goods caused the issues. |

## 4. RESULTS

### *4.1 Baseline Readability*

Across 2,000 articles, vanilla GPT-4o summaries generated without accessibility constraints yielded an average Flesch Reading Ease score well below the target of 90, confirming that standard LLM summarization does not inherently produce dyslexia-friendly output. Table 3 provides a qualitative illustration of the difference between a baseline summary and the output of our iterative pipeline on the same source article.

### *4.2 GPT-4o Iterative Pipeline*

Figure 1 shows the distribution of attempts needed to reach a Flesch Reading Ease score of >= 90 across the dataset. A large proportion of summaries met the threshold on the first attempt (approximately 650 samples), while others required up to four attempts. The distribution is notably bimodal, with peaks at attempts 1 and 3. We attribute this pattern to the dynamics of GPT-4o's corrective refinement: when a summary narrowly misses the readability target on the first pass, the second-attempt correction tends to restructure sentence boundaries in a way that temporarily increases average sentence length before a third, more targeted simplification pass breaks them down further. Essentially, attempt 2 can overshoot in complexity before attempt 3 converges on a simpler form. This behaviour suggests that a smarter stopping criterion, one that detects this oscillation and applies a stronger simplification directive on the second pass rather than a generic corrective prompt, could reduce the proportion of articles requiring three or more attempts.

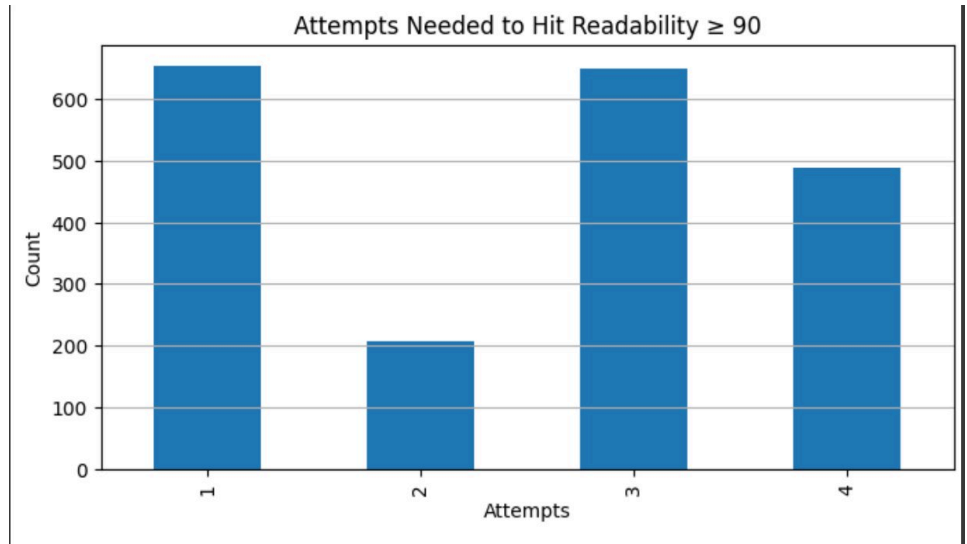

*Figure 1: Distribution of attempts needed to reach Flesch Reading Ease >= 90 across ~2,000 samples. The bimodal pattern (peaks at attempts 1 and 3) suggests two distinct difficulty classes in the dataset.*

Figure 2 shows the final readability distribution after the iterative pipeline. The majority of optimised summaries score between 80 and 105 on the Flesch scale, with the peak bin falling between 90 and 95. This confirms that the pipeline reliably drives outputs into the target readability range for most samples, though a long tail of harder cases scores below 60.

*4.3 Composite Score Across the Dataset*

Figure 3 plots the composite score for the iterative pipeline across all 2,000 articles. Scores range from 0.13 to 0.73 with a mean near 0.55 (SD approx. 0.10). The distribution is stable across the dataset with no systematic degradation over article index, suggesting the pipeline performs consistently regardless of topic or domain. The per-example variance reflects genuine differences in source article complexity: articles with dense, technical language tend to produce lower composite scores even after multiple refinement iterations.

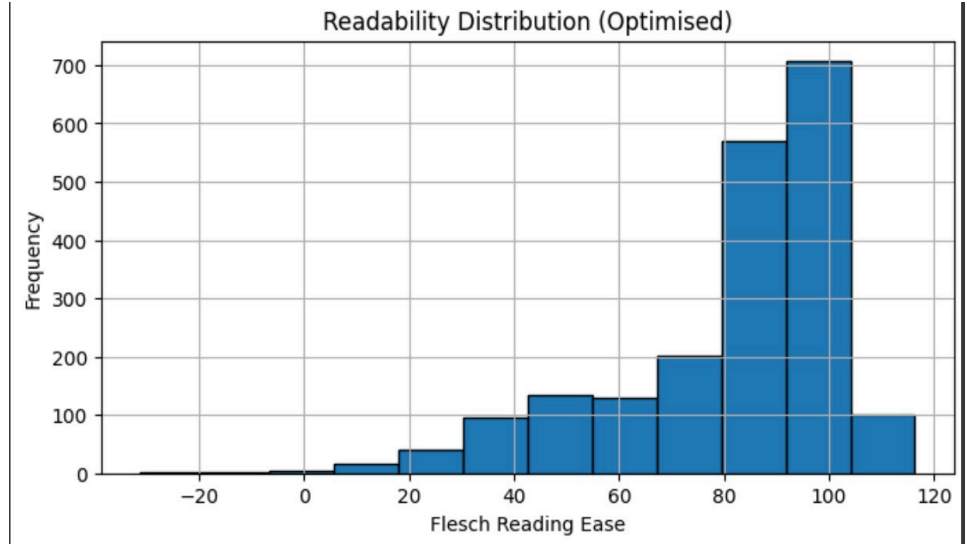

*Figure 2: Final readability distribution (Flesch Reading Ease) across the dataset after pipeline optimisation. The majority of summaries fall in the 80-105 range, with a peak between 90 and 95.*

*4.4 Readability-Fidelity Trade-off*

The composite score data illustrates the inherent tension between readability and semantic fidelity. While the iterative pipeline successfully pushes Flesch scores above the target threshold for most samples, gains in readability do not always correspond to maintained semantic fidelity. Samples requiring three or four attempts tend to show lower composite scores, indicating that aggressive simplification introduces semantic drift. This reinforces the case for monitoring both dimensions jointly and supports the use of a composite stopping criterion rather than optimising for readability alone.

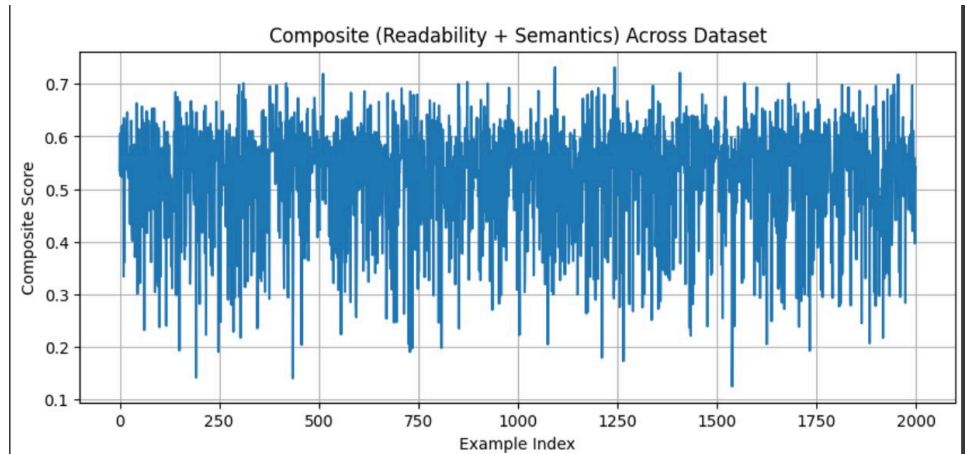

*Figure 3: Composite score (readability + semantic fidelity) across ~2,000 dataset examples. Scores are stable around 0.55 on average, with high per-example variance reflecting differences in source article complexity.*

**5. DISCUSSION**

The results demonstrate a clear and measurable benefit of accessibility-driven prompt engineering over vanilla GPT-4o summarization. The iterative pipeline reliably achieves FRE >= 90 for most samples within four attempts, a result that generic prompting does not achieve. The bimodal attempt distribution is a notable finding: it points to two distinct difficulty classes in the dataset, articles that simplify readily on the first pass, and structurally complex articles that resist simplification initially but respond on a third attempt. Identifying what linguistic features characterise each class is a valuable direction for future work.

The composite score analysis reveals that readability gains do not always come for free semantically. Samples requiring more attempts tend to show lower composite scores, suggesting that aggressive simplification introduces semantic drift. A dynamic stopping criterion, one that halts refinement when the composite score starts declining rather than when readability alone exceeds the threshold, could better balance both objectives.

The composite score analysis shows that two refinement iterations represents the practical optimum: beyond this point, readability gains are marginal while semantic fidelity continues to decline. A dynamic stopping criterion based on the composite score rather than readability alone could further improve pipeline efficiency, halting early when additional iterations are unlikely to improve the joint objective.

## 6. LIMITATIONS

Several limitations constrain the conclusions that can be drawn from this empirical study. First, while the 2,000-sample CNN/Daily Mail corpus is sufficient for prototype-level evaluation of the GPT-4o pipeline, it draws from a single source and domain. Larger and more diverse corpora will be needed for cross-domain generalisability.

Second, the dataset draws exclusively from CNN/Daily Mail news articles, a single source domain. Broader coverage across educational, clinical, and legal texts will be needed to assess whether the prompt engineering strategy generalises to other genres.

Third, the ROUGE and BERTScore metrics in this study are computed against reference summaries written by the authors, who also designed the prompts. This introduces a potential style bias: the pipeline may score favourably not because it produces objectively dyslexia-friendly text by any general standard, but because its outputs resemble the writing style of the prompt designers. Future evaluations should use independently written reference summaries or established dyslexia-friendly corpora to decouple prompt design from metric evaluation.

Fourth, and most fundamentally, all evaluation in this work is automated. Flesch Reading Ease captures sentence length and syllable count but does not model the cognitive processes specific to dyslexia, such as phonological decoding difficulty or working memory load. A summary can score above 85 on FRE while still containing phonologically complex words that are particularly challenging for dyslexic readers. Human-subject evaluation is therefore an essential next step.

Fifth, the study focuses exclusively on news text in English. Transfer to other genres, including textbooks, legal documents, and health information, across other languages remains an open question.

## 7. ETHICAL CONSIDERATIONS

This work is an engineering baseline study and does not constitute a clinical intervention or a validated accessibility tool. The results demonstrate that accessibility-driven prompt engineering measurably improves automated readability scores, but make no claim about cognitive or comprehension outcomes for individuals with dyslexia. The Flesch Reading Ease score is used here as a practical, reproducible proxy for text simplicity, not as a clinically validated measure of dyslexia-specific accessibility. This distinction is important: a summary scoring above 90 on FRE may still contain phonologically complex words or unusual constructions that are challenging for dyslexic readers. Validation with dyslexic users remains essential future work.

This study involves automated evaluation only and does not recruit human subjects; accordingly, no IRB approval was required for the current phase. All planned human-subject studies will seek full IRB approval, implement informed consent, and follow established protocols for research with participants with learning disabilities.

We acknowledge that designing for a heterogeneous population such as people with dyslexia carries a risk of over-generalisation. Dyslexia manifests differently across individuals; a system optimised for average readability gains may not benefit, and could in edge cases hinder, specific users. Future work will incorporate participatory design methods to ensure that the community the system is intended to serve is actively involved in its development.

## 8. FUTURE WORK

We identify four primary directions for extending this research. Human evaluation: IRB-approved comprehension studies with dyslexic readers will provide ground-truth evidence on whether automated readability improvements translate to measurable cognitive benefits. Dynamic stopping: replacing the fixed iteration cap with a composite-score-based stopping criterion could reduce unnecessary refinement passes and semantic drift. Personalisation: adaptive pipelines that tune prompt constraints to individual reading profiles could better serve the diversity within the dyslexic population. Domain expansion: evaluation on educational, clinical, and legal texts will test whether the accessibility-driven prompting strategy generalises beyond news.

## 9. CONCLUSION

This paper presented an empirical study of accessibility-driven prompt engineering for dyslexia-friendly text summarization using GPT-4o. By combining explicit accessibility constraints, few-shot dyslexia-friendly examples, and an iterative corrective feedback loop, our pipeline reliably produces summaries that meet a Flesch Reading Ease target of >= 90, a level that vanilla GPT-4o summarization does not achieve. Evaluated on 2,000 CNN/Daily Mail articles, results show that most summaries meet the threshold within four attempts, with a bimodal distribution suggesting two distinct difficulty classes in the data. A composite score combining readability and semantic fidelity averages 0.55 across the dataset, with per-example variance reflecting the tension between simplification and content preservation. These findings demonstrate that targeted prompt engineering is a practical, training-free strategy for improving LLM accessibility and motivate future human-centered evaluation with dyslexic readers.

## ACKNOWLEDGMENTS

The authors thank the UCARE (Undergraduate Creative Activities and Research Experience) program at the University of Nebraska-Lincoln for supporting model testing and providing feedback throughout this project.